\relax
\documentclass[letterpaper]{article} 
\usepackage{aaai22}  
\usepackage{times}  
\usepackage{helvet}  
\usepackage{courier}  
\usepackage[hyphens]{url}  
\usepackage{graphicx} 
\usepackage{natbib}  
\usepackage{caption} 
\DeclareCaptionStyle{ruled}{labelfont=normalfont,labelsep=colon,strut=off} 
\usepackage{algorithm}
\usepackage{algorithmic}
\usepackage{amsmath}
\usepackage{booktabs}
\usepackage{amssymb}
\usepackage{multirow}
\usepackage{multirow}
\usepackage[dvipsnames]{xcolor}
\usepackage{tabularx}
\usepackage{newfloat}
\usepackage{listings}
\floatstyle{ruled}
\newfloat{listing}{tb}{lst}{}
\floatname{listing}{Listing}
\title{CO2Sum:Contrastive Learning for Factual-Consistent Abstractive Summarization}
\author {
    Wei Liu,\textsuperscript{\rm 1}
    Huanqin Wu,\textsuperscript{\rm 1}
    Wenjing Mu,\textsuperscript{\rm 1}
    Zhen Li,\textsuperscript{\rm 2}
    Tao Chen,\textsuperscript{\rm 1}
    Dan Nie, \textsuperscript{\rm 1}
}
\affiliations {
    \textsuperscript{\rm 1} Tencent AI Platform Department, China\\
    \textsuperscript{\rm 2} Harbin Institute of Technology\\
    \{thinkweeliu,huanqinwu,wenjingmu,vitochen,kathynie\}@tencent.com linklizhen@163.com
}

\usepackage{bibentry}

\DeclareMathOperator*{\argmax}{argmax}

\begin{document}

\maketitle

\begin{abstract}
Generating factual-consistent summaries is a challenging task for abstractive summarization.
Previous works mainly encode factual information or perform post-correct/rank after decoding.
In this paper, we provide a factual-consistent solution from the perspective of contrastive learning, which is a natural extension of previous works. We propose CO2Sum (Contrastive for Consistency), a contrastive learning scheme that can be easily applied on sequence-to-sequence models for factual-consistent abstractive summarization, proving that the model can be fact-aware without modifying the architecture.
CO2Sum applies contrastive learning on the encoder, which can help the model be aware of the factual information contained in the input article, or performs contrastive learning on the decoder, which makes the model to generate factual-correct output summary.
What's more, these two schemes are orthogonal and can be combined to further improve faithfulness.
Comprehensive experiments on public benchmarks demonstrate that CO2Sum improves the faithfulness on large pre-trained language models and reaches competitive results compared to other strong factual-consistent summarization baselines.
\end{abstract}

\section{Introduction}
\begin{figure}[!t]
    \centering
    \centering\includegraphics[width=0.48\textwidth]{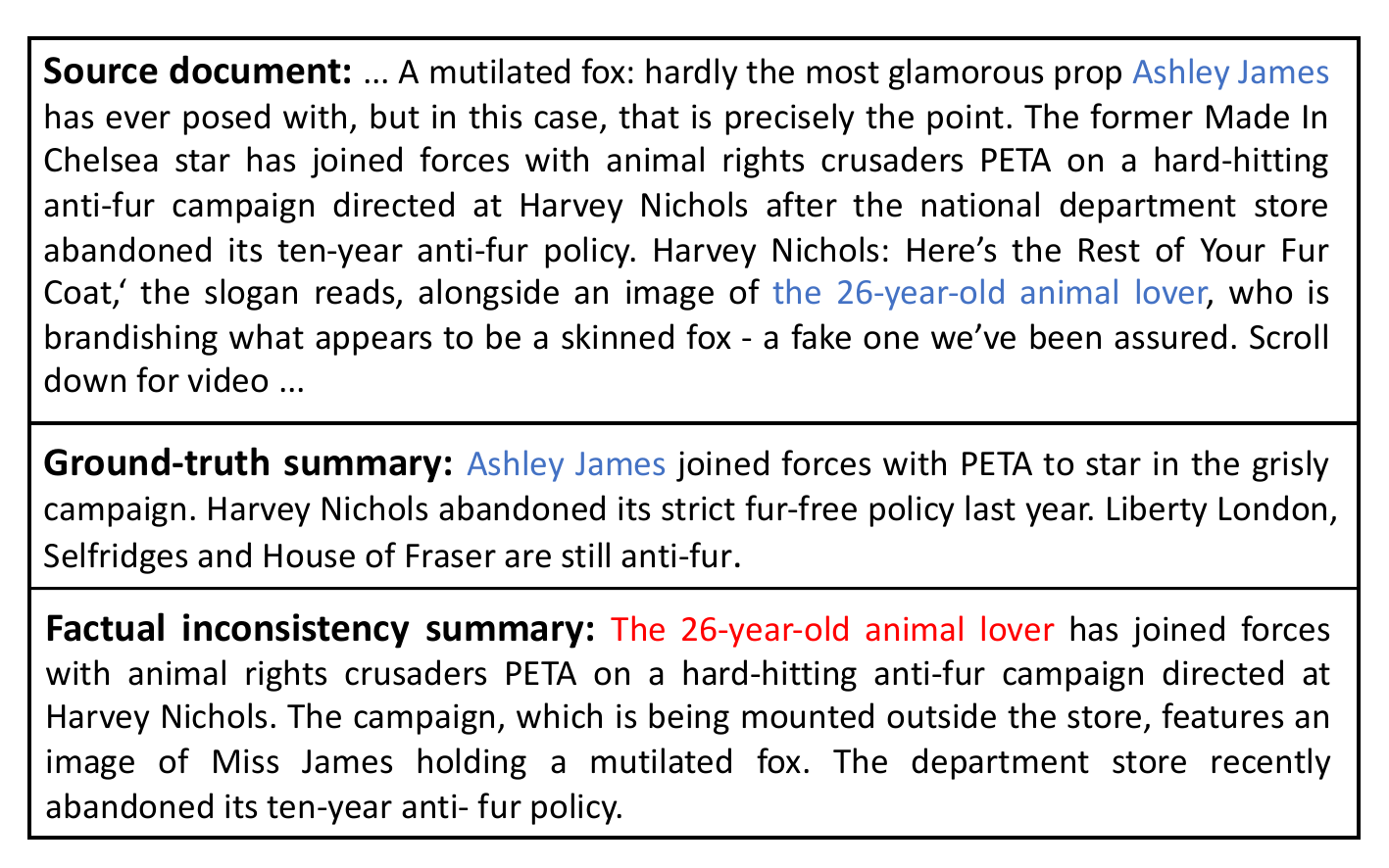}
    \caption{An example of factual-error summary generated by BART. The inconsistent entity is marked with \textbf{{\color{red}red}}.}
    \label{fig:sample_of_factual_error}
\end{figure}

Abstractive summarization aims to generate a concise summary containing core information about the input article.
Recently, large pre-trained language models~\cite{zhang2020pegasus,lewis2020bart} have achieved promising results for generating grammatically correct and fluent summaries and acquired remarkable scores on traditional metrics like ROUGE~\cite{lin2004rouge}.
However, such models are prone to produce summaries with factual-inconsistent errors~\cite{huang2021factual}.
As shown in Figure \ref{fig:sample_of_factual_error}, the model predicts an inconsistent entity ``The 26-year-old animal lover''. It is a correct sentence if no context is given, but the fact is that ``Ashley James'' joined forces not the ``animal lover'', and the ``26-year-old animal lover'' does not refer to ``Ashley James''.

To address such problems in abstractive summarization, several works have been proposed to improve the factual consistency of summarization. As shown in Table \ref{table:method_comparision}, existing works can be roughly categorized into two classes: fact-input methods~\cite{Cao:18, huang-etal-2020-knowledge} which aim to encode the information of facts in the article, and post-edit methods~\cite{cao-etal-2020-factual, chen2021improving} which seek to correct the factual errors after decoding. What's more, there are some integrated works which perform both improvements~\cite{zhu2021enhancing}.
These methods usually need to modify the architecture of the model, adding additional encoding modules or post-edit modules.

Previous works want to model the ``fact-aware'' during encoding and decoding. In this paper, we propose CO2Sum, a novel contrastive learning solution for realizing a fact-aware model.
CO2Sum can improve the faithfulness of summary both on encoding and decoding without introducing extra parameters compared to previous works.
In detail, CO2Sum helps the model to encode fact information in the article, or makes the decoding to be factual correct by distinguishing ground truth summary from the negative summary. It is a natural extension of traditional factual consistency solutions by contrastive learning. Besides, a necessary prerequisite for fact-aware contrastive learning is constructing negative samples containing inconsistent facts. Instead of randomly picking up other summary sentences as negative examples, we propose a language-model-based method for factual inconsistency sample construction.

\begin{table}[!t]
\centering
\small
\begin{tabular}{lcccc}
\toprule[1pt]
Method          & Encoding & Decoding & wo-Params  \\ \hline
Fact-Input     &  \checkmark     &        &          \\
Post-Edit    &       &   \checkmark     &          \\
Integrated  &  \checkmark     &   \checkmark     &          \\
CO2Sum          &  \checkmark    &  \checkmark     &  \checkmark      \\ \hline
\end{tabular}
\caption{Comparison among different methods for factual-consistent abstractive summarization. ``wo-Params'' means the method does not introduce extra parameters}
\label{table:method_comparision}
\end{table}

Experiments conducted on the widely used public datasets validate the effectiveness of our method.
The contributions of this paper can be summarized as follows:
\begin{itemize}
    \item We propose a negative sample construction method LFN (\textbf{L}anguage model-based \textbf{F}actual \textbf{N}egative sample construction). LFN can detect which parts in the summary are easy to produce fact errors and construct negative samples based on it.
    \item We present \textbf{CO2Sum}, a natural extension of previous factual-consistent works in contrastive learning scheme for abstractive summarization. CO2Sum applies contrastive learning directly in the sequence-to-sequence training process without introducing extra parameters.
    \item We validate our method on widely used public datasets. CO2Sum outperforms large pre-trained language model on four factual consistency metrics. The encoding and decoding improvements in CO2Sum are orthogonal and can be further combined to achieve better results.
\end{itemize}
\section{Approach}
\subsection{Overview}
\begin{figure*}[ht]
    \centering
    \centering\includegraphics[width=\textwidth]{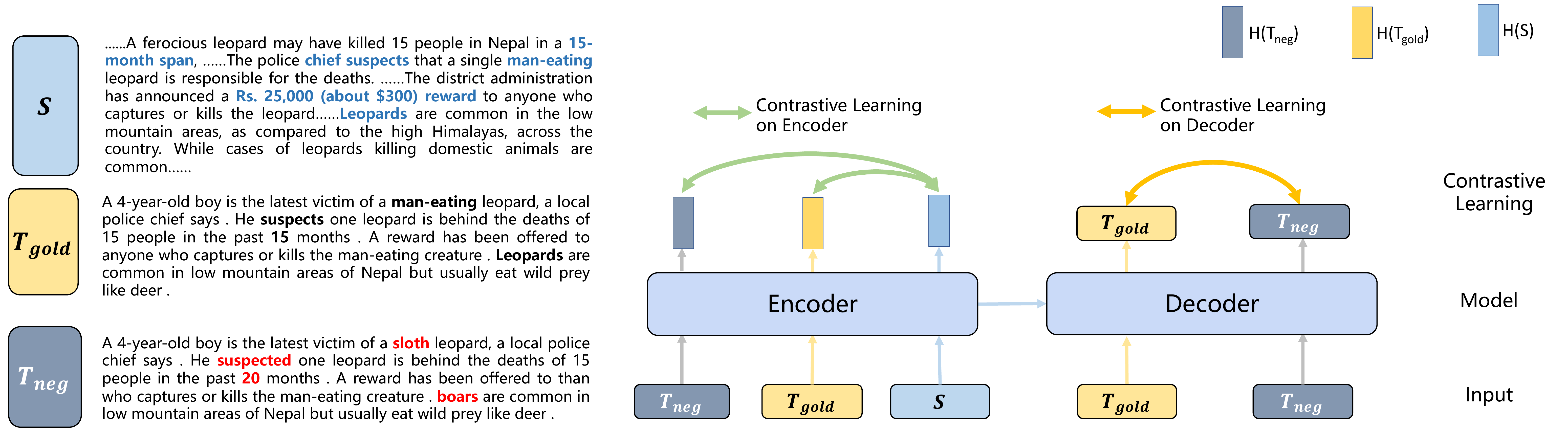}
    \caption{Overview of CO2Sum training process. On the left is a constructed sample $T_{neg}$ along with its ground truth $T_{gold}$ and article $S$ from a real dataset. \textbf{{\color{Cerulean}Article fact}}, \textbf{summary fact} and \textbf{{\color{red}disturbance}} are highlighted. On the right denotes the training process of CO2Sum. We draw target sequence both on the input and output of decoder, which denotes the teacher-forcing training style.}
    \label{fig:overview}
\end{figure*}
Abstractive summarization is a text generation task that predicts the summary of input article word by word. Previous works solve this task with sequence-to-sequence learning.
Most of such models are trained in teacher forcing fashion with maximum likelihood estimation, which means the ground truth label will be given at each time step.
A common problem for such training method is that the model often generates sentences with factual errors, although they are fluent and grammatically correct.
To correct such ``fabrications'' on facts, CO2Sum first construct negative samples which contain inconsistent fact. Then it performs contrastive learning on the encoder and decoder. CO2Sum only changes the process of training and poses no influence on the inference. Figure \ref{fig:overview} gives the overview of CO2Sum. The following sections describe the details of three components in CO2Sum: negative sample construction, contrastive learning on the encoder, and contrastive learning on the decoder.
\begin{algorithm}[ht]
\caption{LFN Algorithm}
\label{alg:LFN}
\begin{algorithmic}[1]
\fontsize{9}{10}\selectfont
\REQUIRE Pretrained language model $LM$ and word embeddings $E$, input article $S$, ground truth summary $T_{gold}$, context $T_{next}$, and the function $V(x)$ to get the vocabulary of $x$
\ENSURE Negative sample summary $T_{neg}$
\STATE initialize $C_0 = T_{gold}; C=\{\}; T_{neg}=T_{gold}$ 
\FOR[T denotes max iteration times] {$i=1;i<=T;i++$} 
    \STATE $C_i$ = $\{\}$ 
    \FOR[L denotes max span length] {$l=1;l<=L;l++$}
        \STATE $SP_l = \{span \in C_{i-1}|length(span)=l\} $
        \STATE $C_d = \{c-span|c \in C_{i-1}, span \in SP_l\}$
        \STATE $C_i = C_i \cup  C_d$
    \ENDFOR
    \STATE $C = C \cup C_i$
\ENDFOR
\STATE $C_{rank} = sorted(C, key=lambda \ x:LM(x))$
\STATE $T_{fragment} = argmin_{c \in C_{rank}[:topk]} LM(T_{next}|c)$
\FOR {$w \in V(T_{fragment})$}
    \STATE $w_{replace} = \argmax_{w_s \in S}Dis(E(w_s), E(x))$
    \STATE $replace \ w \ in \ T_{neg} \ with \  w_{replace}$
\ENDFOR
\RETURN $T_{neg}$
\end{algorithmic}
\end{algorithm}
\subsection{Negative Sample Construction}
It is crucial to build high-quality negative samples for contrastive learning. Negative samples related to factual inconsistency can effectively help the model to be aware of the key facts in the input article, or correct possible fact errors in the output summary. An intuitive way to construct inconsistent negative samples is to replace the entities or noun phrases in the ground truth summary. However, as mentioned in \citet{chen2021improving}, entity faithfulness does not equal to summary faithfulness. It is difficult to cover all facts in the article by hand-craft rules. Intuitively, facts in the sentence are critical for predicting the context. If the facts are deleted or disturbed, the sentence will have less or no relationship with the context. Thus we can identify facts by disturbing them in the sentence and observing the language model probability of predicting the context based on this sentence. This is actually a application of information bottleneck~\cite{west2019bottlesum}. Here we use a pre-trained language model to identify such parts and propose LFN, Language model-based Factual Negative sample construction. Some examples of LFN are shown in the Appendix.

We improve the sentence compression algorithm in the \citet{west2019bottlesum} and apply it to find factual fragments in the ground truth summary, then replace it with the embedding-similar word in the article. Algorithm \ref{alg:LFN} describes the process of LFN:
\begin{itemize}
    \item \textbf{Candidate Generation}: LFN first performs an iterative deletion to generate candidates: it finds spans $SP$ with various length $l$ in the summary $T_{gold}$, then deletes these spans from the summary, generating a series of compressed sentences. These sentences will be used to generate shorter ones in the next iteration. $L$ denotes the maximum length of spans and $T$ denotes the maximum iterative times. Sentences generated from all iterations are used as candidates $C$ for finding factual fragments of the ground truth summary. This step is described in lines 1-10.
    \item \textbf{Candidate Ranking}: All candidates $c \in C$ are then sorted by a two-phase ranking. In the first phase, candidates are ranked by the pre-trained language model probability $LM(c)$, which evaluates the prune score. In the second phrase, top k candidates in the ranked result from the first phase will be re-ranked based on the conditional language model probability of $c$ given context $T_{next}$. We concatenate $T_{next}$ and $c$ to calculate $LM(T_{next}|c)$, which evaluates the relevance score. Through this two-phase ranking for prune and relevance~\cite{west2019bottlesum}, we regard the candidate with the highest score as the factual fragments $T_{fragment}$ of ground truth summary. This step is described in lines 11-12.
    \item \textbf{Word Replacement}: Words in the factual fragments are easy to produce fact errors, so LFN replaces these words in the ground truth summary with embedding-similar (using faiss~\cite{JDH17}) article words to construct negative samples. This step is described in lines 13-17.
\end{itemize}
It is worth noting that the original algorithm in \citet{west2019bottlesum} compresses each sentence in the article, using the next sentence as the context. In LFN's scenario, we compress sentences in the ground truth summary instead of the article. Sentences in summary may not be coherent. Next sentences can not be used as context. So to overcome this problem we find the oracle sentence~\cite{nallapati2017summarunner} in the article for each summary sentence $G$, then use the next article sentence of the oracle as the context $G_{next}$. Such selection of next-to-oracle leads to better coherence than simply picking up the next sentence in summaries. LFN randomly choices one from top k embedding-similar article words to replace the words in the summary for better diversity of negative samples. The reason why LFN finds embedding-similar words from the article not the open vocabulary is that the vocabularies of article and summary in each sample pair tend to be similar, which is ideal for building hard negative samples.

\subsection{Sequence-to-Sequence Learning}
CO2Sum improves on the attention-based sequence-to-sequence learning. Given abstractive summarization dataset with $N$ samples $D = \{S_{i},T_{i}\}_{i=1}^{N}$, where $S$ are input articles and $T$ are output summaries with length $L$. A typical approach for solving such a problem is to leverage the encoder-decoder architecture to model the conditional distribution.
The training loss is cross-entropy defined as:

\begin{small}
\begin{gather}
    P_{s}(x) = \log p(x | S) \\
    L_{CE} = \frac{1}{L} \sum_{j=1}^{L} P_{s}(T_{i,j})
\end{gather}
\end{small}
where $P_{s}$ denotes conditional language model probability.

In this paper, We use the pre-trained sequence-to-sequence model BART~\cite{lewis2020bart} as the baseline architecture, which is a transformer model~\cite{vaswani2017attention} pre-trained on the denoise text generation task.

\subsection{Contrastive Learning on Encoder}

\begin{figure}[!t]
    \centering
    \centering\includegraphics[width=0.47\textwidth]{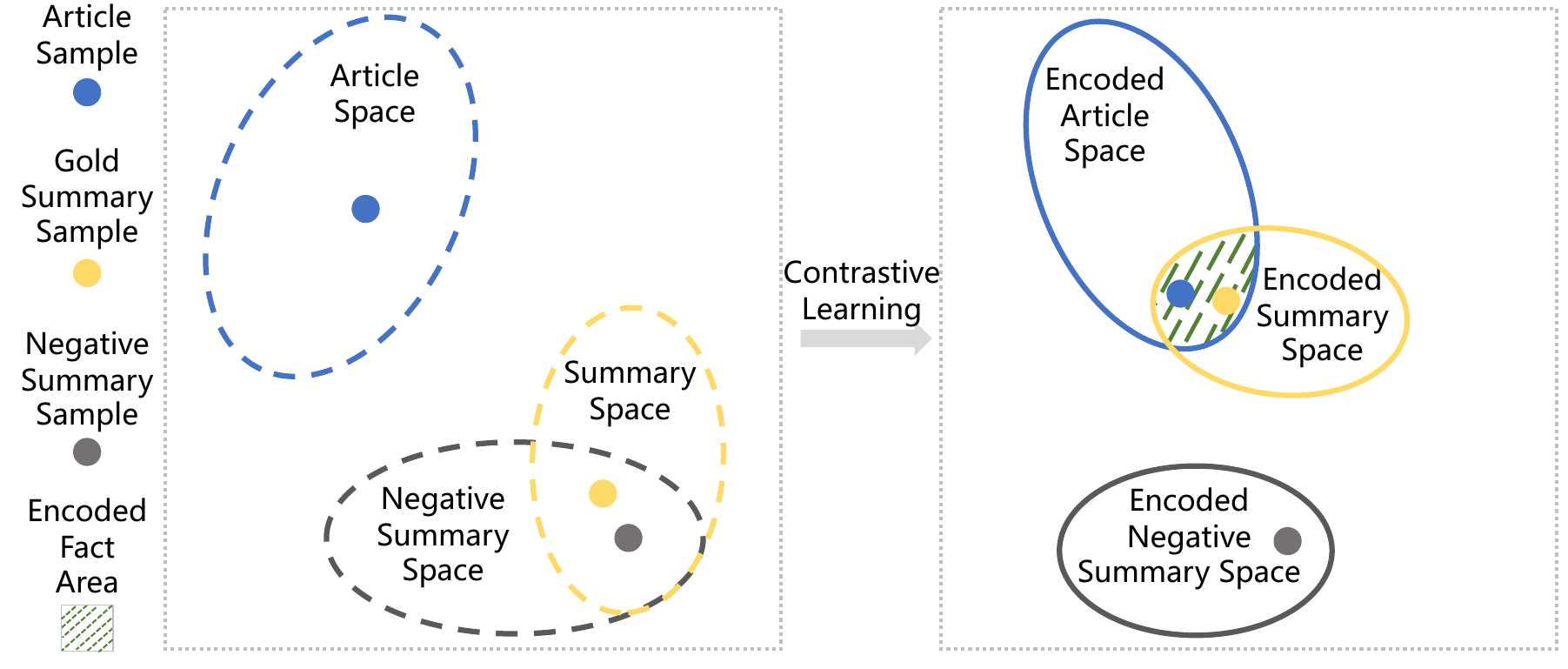}
    \caption{Illustration of contrastive learning on encoder.}
    \label{fig:process-CL-encoder}
\end{figure}

Contrastive learning on encoder (\textbf{CoEnc}) calculates the contrastive loss described in \citet{pan2021contrastive}. CoEnc first encodes the article and summaries (ground truth and negative samples), then make the encoded representation of the article and the ground truth summary closer, and make that of the article and the factual inconsistency summary apart. As shown in Figure \ref{fig:process-CL-encoder}, the motivation of encoding both article and summary on the encoder is to catch and encode fact information. Given the article, the encoder can only distinguish the ground truth summary from the very similar negative summary by catching the common correct fact in the article-summary pair. It can be also explained from the view of data augmentation. Similar to the crops and rotations of images in SimCLR~\cite{chen2020simple}, we can regard the article and summary as two kinds of ``data augmentation'' on the fact. CO2Sum is designed to catch the fact behind the augmentation.

Formally, given a triplet example $(S,T_{gold},T_{neg})$, The objective of contrastive learning on encoder is to minimize the following loss:
\begin{small}
\begin{equation}
\begin{aligned}
L_{Enc} = -\log \frac{exp(H(S) \cdot H(T_{gold}) / \gamma)}{\sum_{i=0}^K exp(H(S) \cdot H(T_{neg_i}) / \gamma)}
\end{aligned}
\end{equation}
\end{small}
where $S,T_{gold},T_{neg}$ denote the input article, the ground truth summary and the negative summary respectively. $H(x)$ denotes the average-pooled encoder output of input text $x$. $K$ denotes the number of negative samples used in each training pair. $\gamma$ is a temperature hyper-parameter, which affects the difficulty of distinguishing positive and negative examples. In general, the higher the value of $\gamma$, the more difficult to distinguish positive examples from negative ones. Intuitively, by maximizing the numerator terms, the loss brings the article and the ground truth summary of relevant facts closer together. Similarly, the article and summary with weak factual consistency are moved away by minimizing the denominator term. Contrastive learning on the encoder allows the model to be aware of the fact in the articles.

\subsection{Contrastive Learning on Decoder}
\begin{figure}[ht]
    \centering
    \centering\includegraphics[width=0.4\textwidth]{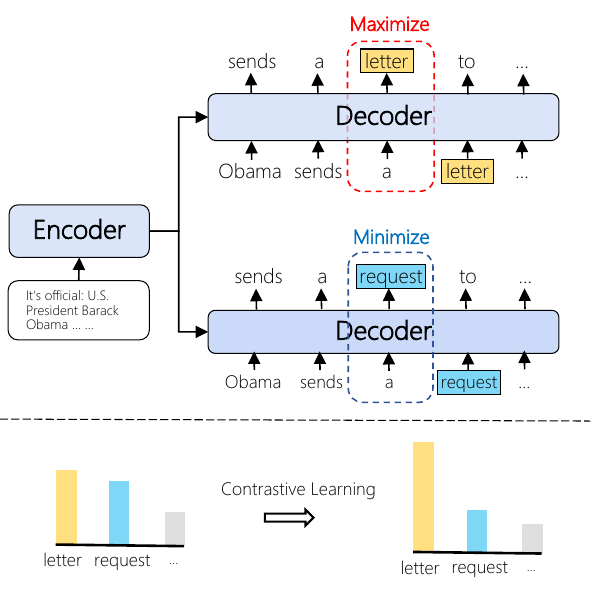}
    \caption{The process of contrastive learning on decoder. The parameters of the two Decoders in the figure are the same. The \textbf{\color{orange}{yellow}} means correct fact, and the \textbf{\color{Cerulean}{blue}} means the inconsistent one. We maximize the probability of ``letter'' and minimize the probability of ``request''.}
    \label{fig:position_no_normal}
\end{figure}
Contrastive learning on decoder (\textbf{CoDec}) is quite different from CoEnc since it does not need the article to be involved in explicitly, and the negative samples are actually ``negative labels''. The object of CoDec is to correct the fact error during decoding. CO2Sum uses max-margin loss~\cite{yang2019reducing} to force the model to increase the decoding probability of ground truth summary while decreases the decoding probability of negative summary.  

The original implementation in \citet{yang2019reducing} uses ground truth and negative summary as labels, respectively. Then it gets two cross-entropy averaged on words to further calculate max-margin loss. However, we found that such implementation will cause instability since most words in the two cases are the same. Optimizing max-margin loss on all word positions will confuse the model. So in CO2Sum, we propose the Position Masked (PM) version of max-margin loss. It just influences the positions where words are replaced by masking other positions. The PM max-margin loss function is like follows:
\begin{small}
\begin{gather}
    L_{Dec} = max\{\frac{1}{|R|} \sum_{i \in R} (P_{s}(T_{neg,i}) - P_{s}(T_{gold,i})) + \eta ,0\} 
\end{gather}
\end{small}
where $R$ means the replaced positions with inconsistent facts. As shown in Figure \ref{fig:position_no_normal}, the word ``letter'' in the third position is replaced as ``request'' in negative sample. There are just tiny differences between these two summaries. The PM max-margin loss only minimizes the probability of inconsistent fact in the replaced position (highlighted in blue) and the probability of words in other positions will not be affected.

\subsection{Model Training}
The encoder contrastive loss or decoder contrastive loss can be simply added to the total loss as a regularization. Also, the two optimizations are orthogonal and can be combined like follows:
\begin{small}
\begin{equation}
\begin{aligned}
    L & =  L_{CE} + \lambda_{Enc} L_{Enc} + \lambda_{Dec} L_{Dec}
\end{aligned}
\label{equ-loss}
\end{equation}
\end{small}
where $\lambda_{Enc}$ and $\lambda_{Dec}$ are coefficients to balance the different training losses. 

\begin{table*}[ht]
\small
\centering
\begin{tabular}{c|c|ccc|cccc}
\toprule[1pt]
\multirow{2}{*}{Dataset} & \multirow{2}{*}{Model} & \multicolumn{3}{c|}{Traditional Metric}          & \multicolumn{4}{c}{Factual-Consistent Metric} \\
                         &                        & ROUGE-1 & ROUGE-2 & \multicolumn{1}{c|}{ROUGE-L} & QAGS   & QuestEval  & Close Fact  & Open Fact \\ \hline
\multirow{4}{*}{CNNDM}   & BART                   & 44.84	& 21.48 &	41.83                      & 70.15 &	30.68 &	54.89 &	41.94     \\
                         & +CoEnc                 &  43.97 &	20.83	 & 40.86                    &  72.28 &	30.67 &	57.05 &	46.52   \\ 
                         & +CoDec                 & 43.73   & 20.71   & 40.68                        & 73.22 &	\textbf{30.79} &	\underline{\textbf{58.19}} &	48.36      \\
                         & +CO2Sum                 &   43.51	 & 20.64	& 40.53                    &  \underline{\textbf{73.87}}	& 30.73	& 58.18 &	\underline{\textbf{49.72}}   \\\hline
\multirow{4}{*}{XSUM}    & BART                   & 43.80 & 20.48 &	34.63 & 13.19  & 15.57 & 2.75 & 2.71  \\
                         & +CoEnc                 & 41.05 &	17.45 &	31.80 & \textbf{13.53} & 16.64 & 2.92	& 3.59   \\
                         & +CoDec                 & 40.84 &	17.23 &	31.56 & 13.27 & 16.73  & 3.03	& 3.31  \\ 
                          & +CO2Sum                 & 40.66 &	17.12 &	31.43 & 13.48 &	\underline{\textbf{16.86}} & \underline{\textbf{3.31}} & \underline{\textbf{4.34}} \\ \bottomrule[1pt]
\end{tabular}
\caption{Results on CNNDM and XSUM datasets. CO2Sum denotes the combination of CoEnc and CoDec. Underlined results denote statistically significantly better ($p < 0.05$).}
\label{table:main_result}
\end{table*}

\section{Experimental Setup}

\subsubsection{Datasets}
We demonstrate the effectiveness of our approach on two public datasets which are widely used by previous factual-consistent works, CNN/Daily Mail (CNNDM) and XSUM. Both datasets collect data from news, containing numerous events, entities, and relations for challenging the factual consistency of summarization models.

\subsubsection{Metrics}
Traditional metrics like ROUGE are limited and perform poorly on evaluating the consistency between article and summary. Following previous works, we evaluate our approach with four factual consistency metrics: 
\begin{itemize}
    \item \textbf{QAGS}~\cite{wang2020asking}: QAGS generates questions about named entities and noun phrases in the predicted summary using a trained QG (Question Generation) model, then uses a QA (Question Answering) model to find answers to questions from the corresponding article. QAGS calculates token-level F1 similarity between QA results and asked entities or noun phrase in the summary as the final score.
    \item \textbf{QuestEval}~\cite{scialom2021questeval}: Compared to QAGS, QuestEval considers the situation of unanswerable questions. What's more, QuestEval does not calculate answer similarity, but scores the precision and recall apart, then QuestEval gives a weighted F1 score.
    \item \textbf{Close Scheme Fact Triple}~\cite{goodrich2019assessing}: Fact Triple based metrics score the precision between summary extracted triple and article extracted triple. The triples $(Subject, Relation, Object)$ are extracted using Named Entity Recognition (NER) and Relation Extraction (RE) models. These triples are structured data of factual information and can be used to evaluate factual consistency.
    \item \textbf{Open Scheme Fact Triple}~\cite{goodrich2019assessing}: Open Scheme Fact Triple is similar to the close ones, but the relationship in the fact triple is text span instead of classified relation label.
\end{itemize}
We use factsumm\footnote{https://github.com/Huffon/factsumm}~\cite{factsumm}, OpenIE\footnote{https://github.com/philipperemy/stanford-openie-python}~\cite{angeli2015leveraging} and official provided code\footnote{https://github.com/ThomasScialom/QuestEval}~\cite{scialom2021questeval} to build evaluation system. On the use of trained models, we choose FLAIR~\cite{akbik2019flair} for NER, LUKE~\cite{yamada2020luke} for RE, T5~\cite{raffel2020exploring} and Roberta~\cite{liu2019roberta} for QA and QG. We calculate all Fact Triple metrics only on oracle sentences in article and summaries, since there is no need to calculate triple precision on those redundant sentences in the article.

\subsubsection{Implementation Details}
For LFN, we use a distilled version of GPT2~\cite{radford2019language} as the scoring language model. The $T$ and $L$ are set to 3. For the experiments of CO2Sum, we use the Fairseq\footnote{https://github.com/pytorch/fairseq} as the implementation of baselines and our method. The pre-trained BART large model is fine-tuned on our training method for 4w steps in CNNDM and 5k steps in XSUM. The maximum number of tokens in a batch is 2048 with gradient accumulation steps of 2. We use Adam optimizer. The $\epsilon$ is 1e-8 and $\beta$ is (0.9, 0.999). The learning rate is set to 3e-5. $K$ is set to 1 in the loss of CoEnc. And the temperature is set to 0.1. We use mixed-precision  to speed up model training. Both $\lambda_{Enc}$ and $\lambda_{Dec}$ are set to 2.0 in Equation \ref{equ-loss} during training. For CNNDM, the warm-up is set to 500 steps. And the warm-up step is set to 125 for XSUM.
All the experiments are done on 16 NVIDIA Tesla V100 GPUs. The training process takes about 12 hours and 3 hours for CNNDM and XSUM.

\section{Results and Analysis}
\subsubsection{Overall Results} 
The performance of summarization on the traditional metric (ROUGE) and factual-consistent metric are shown in Table \ref{table:main_result}. Both CoEnc and CoDec outperform BART on all factual-consistent metrics and all datasets. The combination of CoEnc and CoDec further improves the results. At the same time, similar to other factual-summarization works~\cite{chen2021improving, zhu2021enhancing}, we observe a little drop on ROUGE, which demonstrates that just pursuing ROUGE will conceal the problem of factual inconsistency. All factual consistency metrics on XSUM are much lower than CNNDM. This is because the summaries in XSUM are much more abstractive, and it is difficult for the model to generate consistent results.

\begin{table}[ht]
\centering
\small
\begin{tabular}{c|cccc}
\toprule[1pt]
Methods & QAGS  & QuestEval & Close Fact & Open Fact \\ \hline
Baseline & 70.15 &	30.68 &	54.89 &	41.94 \\
Random                &  70.72	& 30.49	& 55.7	& 42.55    \\
NP                    & 71.18 &	30.59 & 56.99 &	44.16      \\
NER                   & 72.17 &	30.58 &	57.62 &	46.43    \\
LFN              & 72.82 &	30.66 &	57.99 &	47.34  \\
LFN (DN)         & \textbf{73.22}	& \textbf{30.79}	& \textbf{58.19}	& \textbf{48.36}      \\ \bottomrule[1pt]
\end{tabular}
\caption{Comparison between different negative construction methods. LFN denotes our language model based negative sample construction. DN denotes applying Dynamic Negative sample construction.}
\label{table:negative_construction_result}
\end{table}

\subsubsection{Study on Negative Sample Construction}
In this section, we further explore the LFN by conducting different negative sample construction settings.

\begin{table}[ht]
\small
\centering
\resizebox{0.45\textwidth}{!}{
\begin{tabular}{c|cccc}
\toprule[1pt]
\multirow{1}{*}{Replace Ratio}  & QAGS           & QuestEval     & Open Fact        & Close Fact     \\ \hline
Baseline & 70.15 &	30.68 &	54.89 &	41.94 \\
40\%                     & 69.79	& 30.6	& 55.44	& 41.23  \\
30\%                     & 71.43	& 30.63	& 56.86	& 43.55 \\
15\%                     & \textbf{73.22}	& \textbf{30.79}	& \textbf{58.19}	& \textbf{48.36}  \\ \bottomrule[1pt]
\end{tabular}}
\caption{Comparison between different replacement ratios in the negative sample construction.}
\label{table:comparision_ratio}
\end{table}

Firstly, we compare other possible negative construction methods, including:
\begin{itemize}
    \item \textbf{Random}: randomly pick and replace words in the ground truth summary to construct negative samples.
    \item \textbf{NP}: identify noun phrases in the summary and replace the words in the phrase.
    \item \textbf{NER}: perform Named Entity Recognition on the summary and construct negative samples with entity-level replacements.
    \item \textbf{LFN}: the proposed language model based construction method in CO2Sum.
    \item \textbf{LFN (DN)}: similar to the improvement of Roberta~\cite{liu2019roberta} over BERT~\cite{devlin2018bert}, we perform dynamic (DN) negative sample construction during training.
\end{itemize}

As shown in Table \ref{table:negative_construction_result}, LFN outperforms all baselines, proving that only entity or noun phrases can not include all the fact information. The dynamic construction can further improve the result.

In addition, we compare different replacement ratios in the LFN. Higher replacement ratio brings more disturbance. As shown in Table \ref{table:comparision_ratio}, the lowest ratio (15\%) gives the best results. The results of 40\% are similar or worse than the baseline. So hard negative samples with fewer replacements are useful for CoDec. Simple negative samples may harm the original training process since it can be seen as the opposite of label smooth. Only hard negative samples can force the model to identify the detailed fact differences and improve the results.

\subsubsection{Study on CoEnc}
\begin{figure}[ht]
    \centering
    \centering\includegraphics[width=0.4\textwidth]{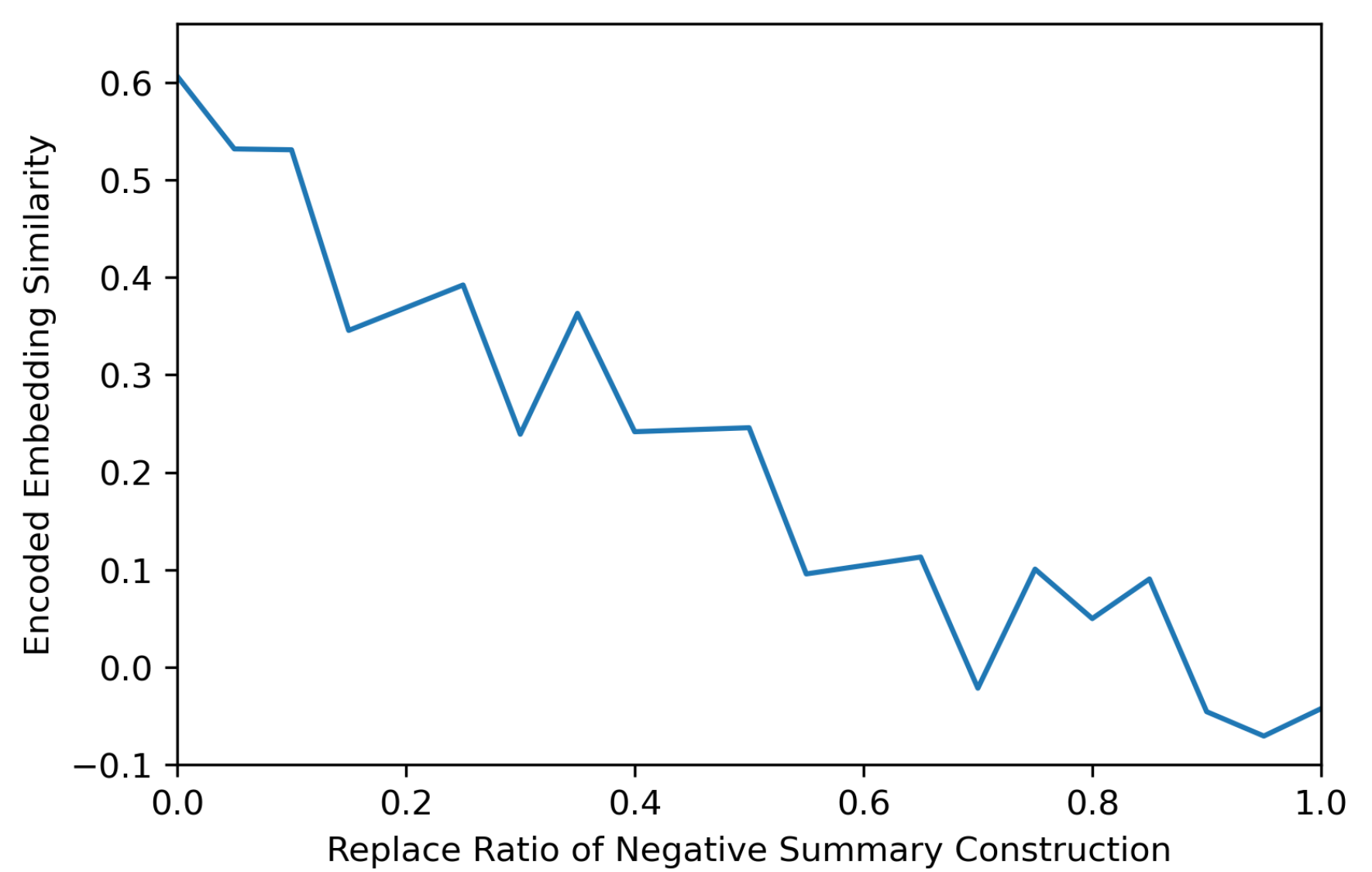}
    \caption{CoEnc as scorer on the fact correctness.}
    \label{fig:study_on_encoder}
\end{figure}
To validate whether the CoEnc is aware of the fact in the article, we use the encoder to score negative summary samples. We first use LFN to construct extra negative samples with different replacement ratios range from 0\% to 100\%. Then we use the encoder as a fact correctness ``scorer'' by calculating the encoded embedding cosine similarity between articles and different summaries. In LFN we replace factual fragments with similar words in the article. Intuitively the summary with higher replacement ratio has more n-gram co-occurrences with article, and the vanilla encoder may give higher scores to it. As shown in Figure~\ref{fig:study_on_encoder}, the encoder assigns higher scores to negative samples with a small replacement ratio (fewer fact errors), and assigns lower scores to those with a larger replacement ratio (more similar to the article but has more disturbance). The encoder effectively catches and encodes the factual fragments in the article and summary, so it can calculate the similarity according to the extent of the fact disturbance. Even though the fact-error samples are more similar to the article, the encoder still gives lower scores.

\subsubsection{Study on CoDec} 
\begin{table}[ht]
\small
\centering
\resizebox{0.45\textwidth}{!}{
\begin{tabular}{c|cccc}
\toprule[1pt]
CoDec  & QAGS           & QuestEval      & Close Fact     & Open Fact      \\ \hline
Baseline & 70.15 &	30.68 &	54.89 &	41.94 \\
Vanilla & 71.57	& 30.56	& 57.02	& 44.42         \\
Gated  & 71.16	& 30.55	& 56.6	& 43.15      \\
PM     & \textbf{73.22}	& \textbf{30.79}	& \textbf{58.19}	& \textbf{48.36} \\
\bottomrule[1pt]
\end{tabular}}
\caption{Comparison on the different loss function of CoDec.}
\label{table:comparision_codec}
\end{table}
In this section, we study the loss function in the CoDec. We compare the results of PM max-margin loss with the original loss (Vanilla) described in \citet{yang2019reducing}. Besides, we attempt another gated-weighting method (Gated) that dynamically calculates the weight of different positions. It uses a Linear Gate Unit~\cite{gehring2017convolutional} to calculate the weights based on the hidden state of the decoder. The results are shown in Table \ref{table:comparision_codec}. The gated method does not perform better than the vanilla, but position masked loss outperforms vanilla on all metrics. We assume that it is too difficult for a model to learn the different weights of positions. A simple mask can stabilize the training and performs better. 

\begin{figure}[ht]
    \centering
    \centering\includegraphics[width=0.4\textwidth]{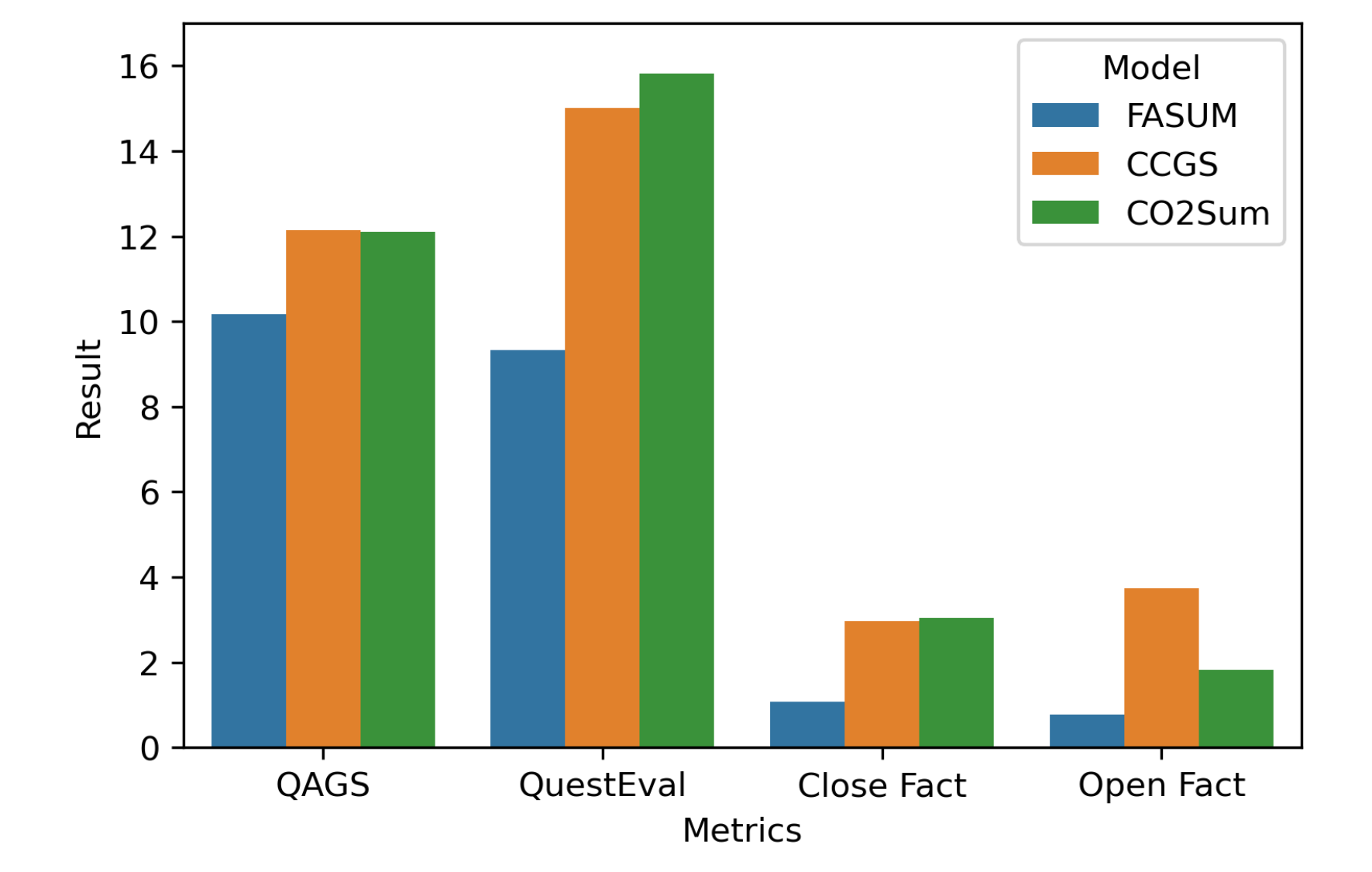}
    \caption{Comparison with other factual-consistent summarization models~\cite{chen2021improving, zhu2021enhancing}.}
    \label{fig:other_paper_compare}
\end{figure}

\begin{figure*}[ht]
    \centering
    \centering\includegraphics[width=\textwidth]{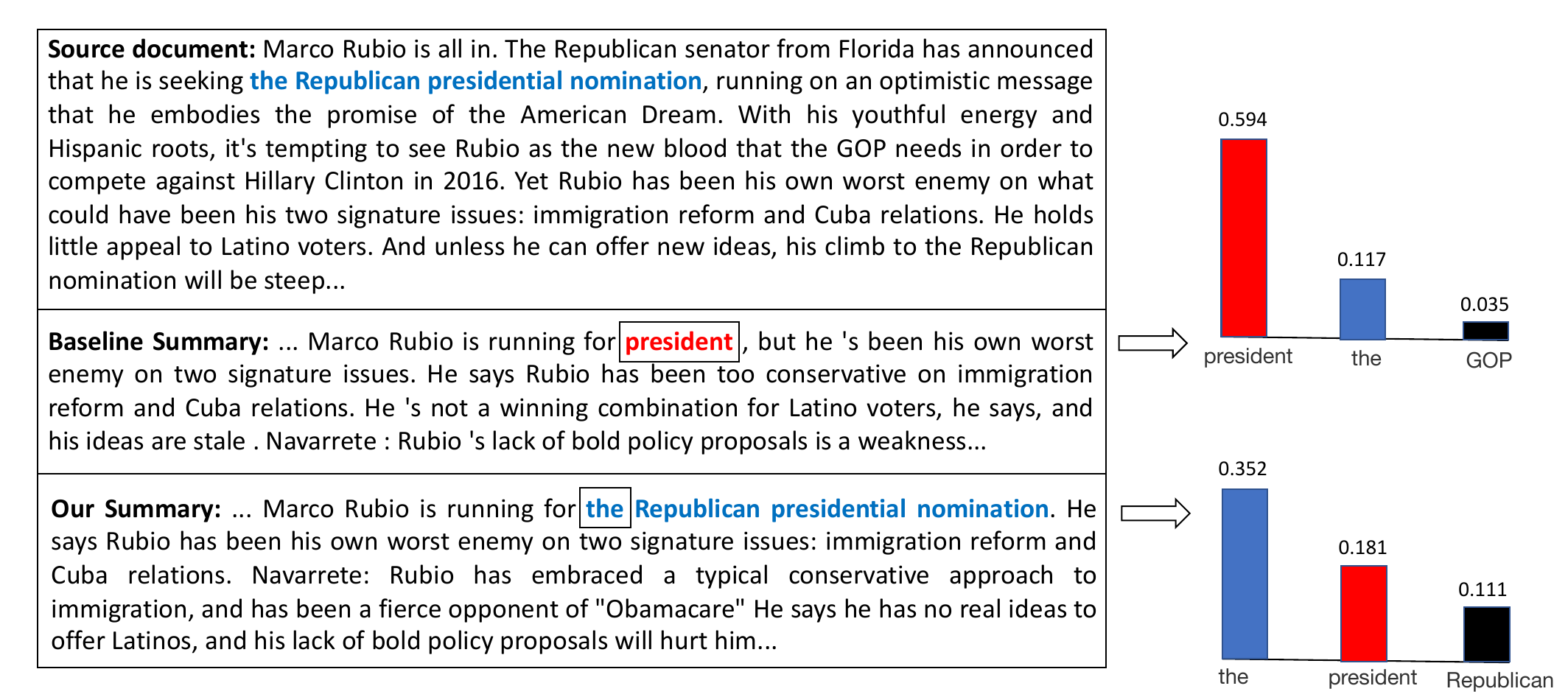}
    \caption{A example of summary generated by BART and our method. The original and inconsistent fact in each summary are highlighted. The decoding probability of the word in the inconsistent position also is shown in right.}
    \label{fig:case_study}
\end{figure*}

\subsubsection{Compare with Previous Works} 
In this section, we compare our model with a strong fact-input baseline FASUM~\cite{zhu2021enhancing} and a strong post-edit baseline CCGS~\cite{chen2021improving}. FASUM uses a knowledge graph to extract fact information and feeds it to the model. CCGS also uses contrastive learning but in the phase of post-ranking. It generates candidates of all possibilities by replacing entities in the decoded results, then CCGS ranks the results by a trained factual-consistent score model. We use the summaries  provided by FASUM\footnote{https://github.com/zcgzcgzcg1/FASum.} and CCGS\footnote{ https://github.com/CogComp/faithful\_summarization/tree/\\master/data.} and evaluated on four factual consistency metrics~\footnote{We only evaluated the 1500 XSum test set since CCGS only provides results on it.}.
As shown in Figure~\ref{fig:other_paper_compare}, our approach consistently outperforms FASUM and achieves competitive results compare with CCGS. It is worth noting that CCGS uses another BART to rank the result, while CO2Sum does not introduce any other extra parameters.

\subsubsection{Case Study}
To further demonstrate the effectiveness of the factual-consistent abstractive summarization method, we give a case study.
We compare the summary generated based on our approach and baseline result base on BART.
As shown in Figure \ref{fig:case_study}, our model can generate factual-consistent summary with phrase ``the Republican presidential nomination'' instead of ``president''.
We further analyze the decoding probability of the word in the inconsistent position. 
Our model can reduce the probability of inconsistent words and increase the probability of correct ones, which confirms the effectiveness of our approach.

\section{Related Work}
\subsection{Contrastive Learning on NLG}
Recently, contrastive learning has been applied to text generation tasks.
\citet{lee2020contrastive} propose to mitigate the exposure bias problem by contrastive learning framework, which maximizes the similarity between positive
pairs and minimizes the similarity between negative pairs.
\citet{liu2021simcls} focus on apply contrastive learning for bridging the gap between the training objective and evaluation metrics.
\citet{yang2019reducing} explore reducing word omission errors in neural machine translation by a contrastive learning approach. 
Compared to these methods, our approach aims to perform factual-consistent abstractive summarization.

\subsection{Fact Consistency for Abstractive Summarization} 
Most existing methods for improving fact consistency can be divided into fact-input-based methods and post-edit-based methods.
Fact-input-based methods focus on enhancing the representation of facts in the source article or incorporating commonsense knowledge, which is useful to facilitate summarization systems understanding the facts for reducing consistent error.
\citet{Cao:18} introduce FTSum to reduce consistent error by applying the encoder to incorporate the fact description.
\citet{li-etal-2018-ensure} aim to incorporate entailment knowledge into the summarization model.
Post-edit based method aims to apply a post-edit on the model-generated summaries for obtaining more factual-consistent summarization.
\citet{dong-etal-2020-multi-fact} propose a fact corrector, which corrects the factual error in the model-generated summary in an iterative and auto-regressive manner.
\citet{cao-etal-2020-factual} propose a neural-based corrector module to address the factual inconsistent issue by identifying and correcting factual errors in generated summaries.
\citet{zhu2021enhancing} explore to model the facts in the source article with knowledge graphs based on a neural network.
\citet{chen2021improving} study contrast candidate generation and selection to correct the extrinsic fact hallucinations in a post-edit manner.
Comparing with the above works, we aim to improve factual consistency through contrastive learning without introducing extra parameters.  

\section{Conclusion and Future Work}
This paper provides a new perspective for factual-consistent summarization and proposes a training scheme named CO2Sum. It makes the encoding and decoding process to be fact-aware during training. Comprehensive experiments on abstractive summarization benchmarks demonstrate the effectiveness of CO2Sum.

The negative sample construction and contrastive learning method on the sequence-to-sequence model can be easily applied to other text generation tasks. What's more, similar to the application of contrastive learning in computer vision unsupervised training, we can extend this method to the pre-training phases of large language models.

\newpage


\nobibliography*

\bibliography{aaai22}

\end{document}